\documentclass{article}

\usepackage{PRIMEarxiv}

\usepackage[utf8]{inputenc} 
\usepackage[T1]{fontenc}    
\usepackage{hyperref}       
\usepackage{url}            
\usepackage{booktabs}       
\usepackage{multirow}       
\usepackage{amsmath,amssymb,amsfonts}
\usepackage{nicefrac}       
\usepackage{microtype}      
\usepackage{fancyhdr}       
\usepackage{graphicx}       
\graphicspath{{./}{media/}}

\newcommand{\rhow}{\rho_{\mathrm{w}}}
\newcommand{\pp}{\,\text{pp}}

\title{Independent Samples, Correlated Variance:\\
A Learnable Cross-View Cue in Path-Traced Stereo Data
}

\author{
  Po-Ting Lin \\
  Independent Researcher \\
  Tainan, Taiwan \\
  \texttt{botimlin@gmail.com} \\
}

\begin{document}

\maketitle

\begin{abstract}
Path-traced synthetic stereo is a primary training substrate for disparity
networks, and the pipelines that consume it assume Monte~Carlo (MC) rendering
noise is independent across the two views. The assumption is correct at the
level it is stated---individual samples---but silent about the object a network
actually sees. We show that the per-pixel MC \emph{variance fields}, though
built from independent samples, are strongly correlated once aligned by the
ground-truth disparity, and that a network can learn to match with that
correlation. Across 20 indoor scenes the warped correlation is $0.754\pm0.016$
against $0.360$ unwarped; it replicates on a second renderer with a different
sampler and sample budget ($0.743\pm0.044$), and a seed-count analysis puts the
population value near $0.85$, making the measurement a lower bound. The effect
is field-level, not sample-level: at the warp correspondence the signed
per-seed residual correlation is $-0.0004$ while the residual envelope
correlates at $0.38$. A capacity-limited siamese probe given nothing but
variance-field patches reaches $78.8\%$ two-alternative forced-choice accuracy
on held-out scenes and falls to chance once the alignment is destroyed; under
the ordinary single-render condition, decorrelation still costs $2.40$
percentage points in a difference-in-differences design, positive in all six
held-out scenes. A real sensor's variance is fixed by its own signal rather
than by transport difficulty, so across views it is redundant with intensity:
the cue is specific to rendered data.
\end{abstract}


\section{Introduction}\label{sec:intro}

Disparity networks are trained on synthetic stereo because ground-truth
disparity is free at render time. How that stereo is rendered, however, has
changed. The corpora on which the field's stereo networks were
pre-trained~\cite{chang2018psmnet,lipson2021raft,xu2023igev}---%
SceneFlow~\cite{mayer2016large} and Sintel~\cite{butler2012sintel}---were
produced with Blender's internal, non-path-traced engine; Mayer \emph{et al.}
state that they ``modified the pipeline of Blender's internal render
engine''~\cite{mayer2016large}. What is being built now is different in kind:
Hypersim~\cite{roberts2021hypersim} uses a physically based production
renderer, Spring~\cite{mehl2023spring} uses Blender Cycles, and current work
on synthetic-stereo design~\cite{yan2025whatmakes} builds on procedural
path-traced generation~\cite{raistrick2023infinigen}.

That shift adds an ingredient the earlier pipelines did not have: residual
Monte~Carlo (MC) noise. It is treated as an additive nuisance whose two views
are statistically independent.

That statement is true, and weaker than it sounds. The random number streams
driving the two cameras are seeded separately, so \emph{individual samples}
are uncorrelated by construction. But a network does not consume samples; it
consumes images, and the cues it forms are functions of aggregated
intensities. The natural aggregate is the per-pixel variance field
$\sigma^2(x,y)$ across independent renders---an object determined by the
rendering integrand rather than by any particular draw.

\begin{figure}[t]
\centering
\includegraphics[width=\textwidth]{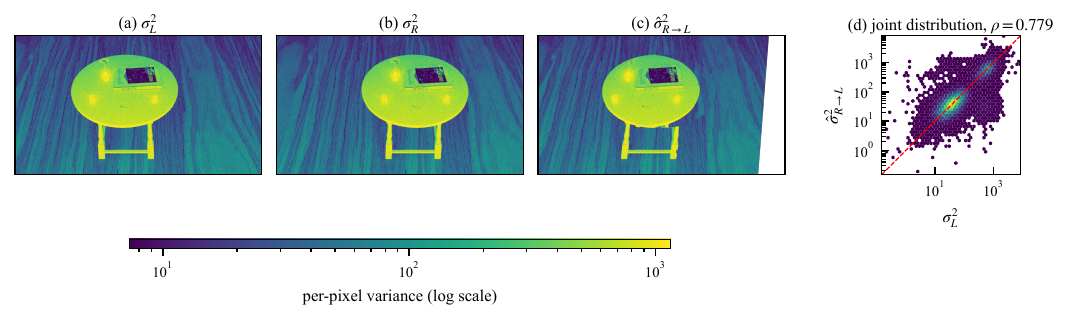}
\caption{The structure, on one scene. (a),~(b) per-pixel MC variance fields
$\sigma_L^2$ and $\sigma_R^2$ from $N{=}30$ independent seeds at $512$\,spp.
(c)~the right field warped into left coordinates by the ground-truth
disparity. (d)~joint distribution of (a) against (c) over the valid set,
with $y{=}x$ in red. The two views' samples are independent by construction;
their variance fields are not, once the geometric transform is applied.}
\label{fig:teaser}
\end{figure}

This paper reports that the two views' variance fields are strongly
correlated once aligned by the ground-truth disparity (Fig.~\ref{fig:teaser}),
and that the correlation is usable: a network trained on nothing else learns
to match with it. Two clarifications are worth making immediately, because
both invert what one might expect.

\textbf{This is about alignment, not magnitude.} The claim is not that
Lambertian surfaces have similar variance in the two views, but that the
\emph{spatial structure} of the two fields coincides after the disparity warp.
The two are in fact anti-correlated across materials: glass carries roughly
seven times the variance of a Lambertian background yet only $0.30$ of the
alignment, against $0.78$ for non-glass.

\textbf{Sample-level independence is preserved, not contradicted.} We measure
it directly. At the warp correspondence the signed per-seed residual
correlation is $-0.0004$; the correlation lives in the residual
\emph{envelope}, which is a property of the integrand and not of the draw.

\textbf{Why this matters to current data generation.} Render quality is
already an engineering and experimental choice in synthetic-stereo generation,
and it measurably changes zero-shot accuracy: Yan \emph{et
al.}~\cite{yan2025whatmakes} ablate it and find noise helps on KITTI and hurts
on cleaner benchmarks, which they attribute to augmentation. Analyses of that
kind read the choice through the per-view \emph{marginal}---how much noise each
image carries. What we report is that wherever the finite-sample MC residual
survives into the released pair, the same choice also governs the presence and
the observability of a disparity-aligned cross-view statistic
(Sec.~\ref{sec:field} separates those two: the normalised alignment does not
decay over the tested budget range, even though the residual magnitude does).
Rendering quality is therefore not only a fidelity parameter; it alters what
matching information the training distribution contains. That is a claim about
data generation as it is practised today, and it requires no deployed network
to be affected.

We are equally explicit about vocabulary. The cue is specific to the rendering
process and demonstrably usable in training-like conditions. We do not measure
whether relying on it helps or harms real-data accuracy, and therefore make no
shortcut claim; we call it a rendering-specific cue.

\noindent\textbf{Contributions.}
\begin{itemize}
  \item We quantify the correlation, show it is carried by the disparity
        alignment rather than by generic field similarity, and replicate it on
        a second, independently configured renderer (Sec.~\ref{sec:struct}).
  \item We separate the field-level effect from the sample-level assumption by
        direct measurement, and show the reported value is a lower bound on
        the population one (Sec.~\ref{sec:field}).
  \item We show the cue is \emph{learnable}: a controlled probe trained only
        on variance fields matches at $78.8\%$ on held-out scenes and returns
        to chance when the alignment is removed, with a
        difference-in-differences design and a placebo control
        (Sec.~\ref{sec:learn}).
\end{itemize}

\section{Related Work}\label{sec:related}

Four literatures border this phenomenon; none covers it, and the ex-ante
position of each is independence.

\textbf{Monte Carlo variance.} Graphics treats $\sigma^2$ as a per-image
convergence quantity---the rate at which one estimator approaches its
mean~\cite{veach1997robust,pharr2023pbrt}---and adaptive sampling and learned
denoising inherit that framing~\cite{zwicker2015recent}. The closest
operational work is Philippi \emph{et al.}~\cite{philippi2023practical}, who
use per-eye variances to set an MSE-optimal stereoscopic blending weight for
real-time ray tracing. Their derivation \emph{explicitly models the per-sample
streams as independent across views}, and estimates each $\sigma^2$ from a
$3\times3$ neighbourhood inside a single render. The relational object we
study---across-seed variance fields correlated under the disparity warp---is
not characterised; the independence they assume is exactly the statement we
preserve.

\textbf{Stereo matching.} The standard texts model the two views as clean
images related by geometry plus independent per-view measurement
noise~\cite{szeliski2022computer,hartley2004multiple}, and the classical
taxonomy~\cite{scharstein2002taxonomy} and cost-volume
networks~\cite{chang2018psmnet,lipson2021raft,xu2023igev} inherit it. Under
that model any cross-view structure beyond geometry does not exist, so a
deterministic cross-view residual signal has no place to sit.

\textbf{Synthetic stereo data.} Published characterisations of these
corpora~\cite{mayer2016large,butler2012sintel,roberts2021hypersim,%
mehl2023spring} describe image-domain quantities---texture, disparity range,
photometric realism---and treat the renderer as a black box that emits clean
pairs. Work sweeping the generator's design
space~\cite{yan2025whatmakes,raistrick2023infinigen} does open that box, but
reports rendering parameters against downstream accuracy; the statistics of
the residual those parameters leave behind, and in particular its behaviour
across the two views, are not characterised. The generational split above is
itself part of why: for most of the period in which stereo networks were
developed, the training corpora had no MC residual to characterise.

\textbf{Shortcut learning.} Chuah \emph{et al.}~\cite{chuah2021shortcut} frame
stereo generalisation failure as shortcut learning~\cite{geirhos2020shortcut}
and name two shortcuts in synthetic stereo: identical local RGB statistics
between matching pixels, and unrealistic synthetic textures. Both are
training-only regularities, and both live at the intensity level; what we
report is the same kind of object one level down, in the rendering residual.
They demonstrate the consequence we stop short of demonstrating.

The effect sits in the seam. It is derivable from the variance integrand
decomposition (Sec.~\ref{sec:disc}) once one thinks to look, but derivability
in hindsight is a property of every true result; the question is whether it
was reported or anticipated, and the recorded position of the nearest
work~\cite{philippi2023practical} is the opposite one.

\section{Method}\label{sec:method}

\subsection{Variance fields and alignment}
Each scene is rendered $N$ times per camera with independent RNG seeds. With
$I_V^{(n)}$ the $n$-th radiance estimate of view $V\in\{L,R\}$ and $\bar I_V$
the across-seed mean,
\begin{equation}
\sigma_V^2(x,y)=\frac1N\sum_{n=1}^{N}\bigl(I_V^{(n)}(x,y)-\bar I_V(x,y)\bigr)^2 ,
\label{eq:var}
\end{equation}
averaged over RGB. Given ground-truth disparity $d$ we warp the right field
into left coordinates,
\begin{equation}
\hat\sigma_{R\to L}^2(x,y)=\sigma_R^2\bigl(x+d(x,y),y\bigr),
\label{eq:warp}
\end{equation}
by bilinear interpolation along $x$; the convention $x_R = x_L + d$ is the one
verified for this renderer, and pixels whose correspondence leaves the image,
or whose $d$ is non-positive or non-finite, are masked out to leave a valid
set $\Omega$. Equation~\eqref{eq:warp} is the alignment every cost-volume
matcher applies implicitly, so any cross-view feature such a network consumes
has passed through it.

We report the Pearson coefficient
$\rhow=\mathrm{corr}(\sigma_L^2,\hat\sigma_{R\to L}^2)|_\Omega$. Pearson is
scale-invariant in both arguments, which matters here because
$\sigma^2\propto1/\mathrm{spp}$~\cite{veach1997robust}: a magnitude-sensitive
metric would confuse a change of sample budget with a change of structure.

\subsection{Finite-seed attenuation}
Both fields in \eqref{eq:var} are estimates, each carrying independent
estimation noise that shrinks with $N$. The measured correlation is therefore
attenuated, and the classical dilution form
\begin{equation}
\rhow(N)=\rho_\infty\,\frac{N}{N+c}
\label{eq:atten}
\end{equation}
lets the population value $\rho_\infty$ be read off a seed-count sweep. We use
this in Sec.~\ref{sec:field} to bound the reported figure from below rather
than to claim convergence of $\sigma^2$ itself, which under heavy-tailed
transport need not exist.

\section{The Cross-View Variance Structure}\label{sec:struct}

\subsection{Setup}
The primary corpus is 20 indoor scenes rendered with Mitsuba~3
\cite{jakob2022mitsuba3} at $1280\times720$, stereo baseline $26$\,mm,
horizontal FOV $45.4^\circ$, path depth 32, $\mathrm{spp}=512$, with $N=30$
independent seeds per camera. Ground-truth disparity and a per-pixel material
mask with a dedicated glass channel come from the renderer.

For the cross-renderer test, five of the same scenes were re-rendered in
Blender~4.2.3~LTS Cycles~\cite{blender2024} from the same geometry. Camera intrinsics
and extrinsics, materials, lighting and bounce budget were aligned item by
item; denoising, adaptive sampling and firefly clamping were disabled, and
output is linear float32. The only deliberate difference is the sample budget
($128$ instead of $512$), which is admissible because $\rhow$ is
scale-invariant. That the two ports describe the same physical scenes is
checked independently: material-mask IoU is $1.0000$ on all five scenes and
the disparity fields agree to $\le 0.39$\,px. The comparison is therefore
paired rather than two independent samples, which is what lets five scenes
carry a replication claim.

\textbf{On the size of the corpus.} Twenty scenes is small, so it is worth
being exact about what it supports. The estimand is a per-scene statistic with
little scene-to-scene variation: for non-glass the standard deviation across
the 20 scenes is $0.011$, giving a standard error of $0.0025$ and a $95\%$
interval of $[0.774,0.784]$. The separation it must resolve---the gap of
$0.419$ between warped and unwarped---is about fifty standard errors wide, so
adding scenes would tighten an interval already two orders of magnitude
narrower than the effect. The learnability results are held to the same
discipline: every interval is a scene-level bootstrap over the six held-out
scenes, never a pixel count, and the headline difference-in-differences rests
on a sign test using nothing but those six ($p\approx0.016$).

\subsection{Correlation across scenes and renderers}
Table~\ref{tab:main} gives the result. On Mitsuba the warped correlation is
$\rhow = 0.754\pm0.016$ over 20 scenes, a coefficient of variation of
$2.1\%$; correlating the same two fields without the warp gives $0.360$. The
alignment, not any generic resemblance between the fields, is what carries the
structure.

The effect survives a change of renderer. Cycles---a different sampler, a
different light-transport implementation and a quarter of the sample
budget---gives $\rhow = 0.743\pm0.044$ over the five ported scenes against an
unwarped baseline of $0.159$, a $4.7\times$ gain from alignment, with the
material ordering preserved in $5/5$ scenes. The two headline means differ by
$0.011$, comfortably inside the sampling error of a five-scene estimate
($\mathrm{SE}=0.020$); the backends are not distinguishable at this corpus
size, which is the claim being made---replication, not a measured difference
between renderers. This is what the integrand argument of
Sec.~\ref{sec:disc} predicts: the structure is a property of MC path tracing,
not of one implementation.

\begin{table}[t]
\caption{Warped correlation $\rhow$, by renderer and material.
Mitsuba: 20 scenes at $512$\,spp. Cycles: five of the same scenes at
$128$\,spp. Both $N=30$ seeds per camera.}
\label{tab:main}
\centering
\begin{tabular}{llcc}
\toprule
& Region & Mitsuba (20) & Cycles (5) \\
\midrule
\multirow{3}{*}{Warped}
  & all pixels & $0.754\pm0.016$ & $0.743\pm0.044$ \\
  & non-glass  & $0.779\pm0.011$ & $0.726\pm0.034$ \\
  & glass      & $0.301\pm0.102$ & $0.487\pm0.079$ \\
\midrule
Unwarped & all pixels & $0.360\pm0.036$ & $0.159\pm0.061$ \\
\bottomrule
\end{tabular}
\end{table}

\subsection{Material dependence}
The ordering is counter-intuitive and stable. Non-glass correlates more than
twice as strongly as glass (Table~\ref{tab:main}) and is an order of magnitude
more consistent across scenes ($1.4\%$ relative variation against $33.9\%$).
Yet glass dominates the absolute noise---roughly $7\times$ the mean variance
of non-glass at $512$\,spp. Cross-view structure is thus \emph{inversely}
related to variance magnitude. We return to why in Sec.~\ref{sec:disc}.

\section{Field-Level, Not Sample-Level}\label{sec:field}

\subsection{The independence assumption, measured}
The dichotomy in the title is not rhetorical, and it is checkable. Taking
warp-corresponding pixel pairs and correlating the two views' per-seed
residuals directly gives a signed correlation of $-0.0004$: sample-level
independence holds to one part in $10^{4}$ at exactly the location where the
cue is claimed. Correlating the residual \emph{magnitudes} at the same
locations gives $0.3797$, and this envelope correlation is insensitive to how
seeds are paired across views ($0.3805$ for identical stream indices,
$0.3816$ two apart). An envelope that does not care which draw produced it is
a property of the integrand, which is precisely what
Eq.~\eqref{eq:var} aggregates.

\subsection{The reported value is a lower bound}
\begin{figure}[t]
\centering
\includegraphics[width=0.8\linewidth]{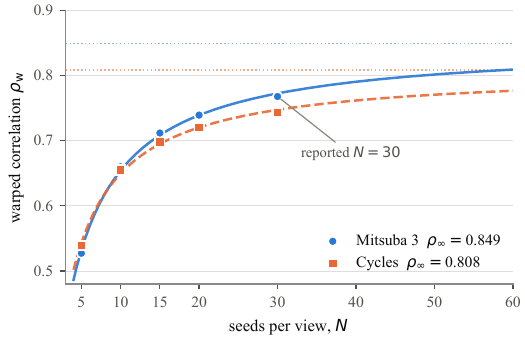}
\caption{Seed-count attenuation, cross-scene means over the five shared
scenes. Markers are measured; curves are the fit of Eq.~\eqref{eq:atten},
extrapolated, with the asymptote dotted. The two backends track each other to
better than $0.025$ at every budget, and $N{=}30$---the value reported
throughout this paper---sits well below the asymptote.}
\label{fig:atten}
\end{figure}

Sub-sampling the same renders at $N\in\{5,10,15,20,30\}$ traces the
attenuation curve of Eq.~\eqref{eq:atten} (Fig.~\ref{fig:atten}). Two things
follow. First, the curve is a statistical property rather than a renderer
property: the two backends agree at every budget to better than $0.025$.
Second, $N=30$ has not saturated. Fitting \eqref{eq:atten} gives
$\rho_\infty = 0.849$ ($c=2.98$, maximum residual $0.006$) for Mitsuba and
$\rho_\infty = 0.808$ ($c=2.44$, maximum residual $0.005$) for Cycles, so the
$0.754$ reported above understates the underlying correlation.

A sweep over sample budgets agrees: the non-glass correlation stays within
$0.785$--$0.788$ across a $16\times$ range of spp, while the residual
magnitude falls by that factor. The effect is not a low-budget artefact---
though what persists is the normalised structure, not necessarily how easily a
network reaches it, and Sec.~\ref{sec:learn} shows the two can come apart.

\section{Is the Cue Learnable?}\label{sec:learn}

A correlation in the data is not yet a cue in a matcher. This section asks
whether a network can learn to match \emph{with} it.

\subsection{Probe design}
We deliberately use a small, purpose-built probe rather than a full
disparity network. A capacity-limited comparator with a discriminative
read-out isolates the question---is this signal usable for correspondence---
from confounds that a large architecture would introduce on a 20-scene
corpus, and controlled probes of this kind are the established instrument in
the shortcut-learning literature~\cite{geirhos2020shortcut,chuah2021shortcut}.
It measures what the data makes available to a matcher of the class that
cost-volume architectures embed.

The probe is a shared-weight siamese encoder---five valid $3\times3$
convolutions with GroupNorm, scoring by inner product of $L_2$-normalised
embeddings, $139{,}233$ parameters, no disparity regression. It sees
$13\times13$ patches. The 20 scenes are split $14/6$, the split frozen before
any training run; all numbers below are on the six held-out scenes with
scene-level bootstrap confidence intervals over $10^{4}$ resamples. Training
is a hinge ranking loss for a fixed $20{,}000$ steps, identical across
conditions. Read-out is two-alternative forced choice (2AFC) between the true
correspondence and a decoy at offset $\delta\in\pm\{1,2,4,8\}$; chance is
$50\%$.

Three data conditions are used. \textbf{Normal} is the rendered pair.
\textbf{Decorr} replaces the right residual by a $16\times16$ block-shuffled
copy, $\tilde I_R^{(n)} = \bar I_R + \Pi(\epsilon_R^{(n)})$, which preserves
the clean image and first-order intensity statistics and destroys only the
cross-view alignment. \textbf{Placebo} permutes which seed's residual is
paired with which, leaving the variance field untouched; it should be
indistinguishable from Normal, and is the control that separates the cue from
any side-effect of perturbing residuals.

\subsection{Variance fields alone are enough}
Given $(\sigma_L^2,\sigma_R^2)$ patches and no intensity whatsoever, the probe
reaches $78.77\%$ and $78.82\%$ on the two training seeds
(Table~\ref{tab:learn}), against a $50\%$ chance floor. Destroying the
alignment at test time collapses the same model to $50.17\%$ and $49.80\%$,
and models trained on decorrelated data never rise above chance during
training---their loss sits at the hinge margin, meaning no exploitable
gradient signal exists once alignment is gone. The signal concentrates in
non-glass ($79.25\%$) with glass barely above chance ($52.63\%$), mirroring
$\rho_{\neg\text{glass}}=0.779$ against $\rho_{\text{glass}}=0.301$.

A classifier two-sample test agrees and adds the placebo. Normal and
decorrelated pairs separate at $\mathrm{AUC}=0.893$ $[0.886,0.903]$; the
seed-permutation placebo---which leaves the variance field intact---is
indistinguishable at $0.503$, its loss converging to $\ln 2$. A seamless
$16$\,px shift, which produces no block boundaries, is still detectable at
$0.690$, so the $0.893$ is not reading shuffle seams.

\begin{table}[t]
\caption{The input ladder. 2AFC accuracy on six held-out scenes, chance
$=50\%$; ``acquired'' counts initialisations that leave the hinge margin
within the fixed step budget, and the accuracy column reports those that do.
Every acquirer collapses to chance once the cross-view alignment is
destroyed.}
\label{tab:learn}
\centering
\setlength{\tabcolsep}{3.5pt}
\begin{tabular}{lccccc}
\toprule
Input & $N$ & Params & Acq. & Normal & Decorr \\
\midrule
$\sigma^2$ fields          & 30 & 139K & $2/2$ & $0.788$, $0.788$ & $0.502$, $0.498$ \\
residuals, perm.-inv.      & 30 & 144K & $1/4$ & $0.692$          & $0.499$ \\
residual                   & 1  & 139K & $1/6$ & $0.548$          & $0.500$ \\
residual                   & 1  & 1.03M & $0/3$ & ---             & --- \\
\bottomrule
\end{tabular}
\end{table}

\subsection{The network can build the statistic itself}
Handing the probe a precomputed $\sigma^2$ is a favourable setting. Given
instead the raw $30$-seed residual stack---no variance anywhere in the
input---and forced to be invariant to seed order, it still reaches $69.24\%$
and transfers unchanged across a seed permutation ($69.18\%$): the network
constructs the variance statistic implicitly, from data in which it is only
latent.

What the ladder of Table~\ref{tab:learn} adds is where this breaks. Acquisition
grows fragile as the seed budget falls---$2/2$, then $1/4$, then $1/6$---and at
$N{=}1$ the surviving initialisation reaches only $54.82\%$
$[54.63,54.98]$, concentrated entirely in non-glass ($54.93\%$, against glass
at chance). Raising capacity sevenfold does not rescue it, so the binding
constraint is the signal-to-noise ratio of a single-render envelope rather than
model size.

\subsection{The realistic condition}
The conditions above isolate the cue by withholding intensity. The practical
question is whether it is used when photometric evidence is available, which
is the ordinary training regime: one render per view, full images. Here a
plain before/after comparison is not enough, because decorrelated data might
simply be harder. We therefore use a difference-in-differences design,
subtracting the degradation of a control model trained on decorrelated data
from that of the normally-trained model.

Normally-trained models lose $2.98\pp$ and $3.61\pp$ when tested on
decorrelated data; the controls lose $0.88\pp$ and $0.91\pp$. The net
difference-in-differences is $2.40\pp$. It is positive in all six held-out
scenes, spanning $2.07$--$2.74\pp$---a one-sided sign test gives
$p = 2^{-6}\approx0.016$---and in all four runs, with a scene-level bootstrap
CI of $[2.21,2.58]\pp$. The reliance is modest in size, as one should expect
when full photometric evidence is available; what matters is that it is not
zero. The measurement establishes that the joint structure of the residual
reaches learned matching behaviour under the condition ordinary training
operates in---not that it dominates that behaviour.

The comparison that gives it meaning is the same design applied to the
$N{=}30$ condition, where averaging thirty seeds has already suppressed most of
the photometric noise: there the net figure is only $0.19\pp$, likewise
positive in all six scenes. Reliance grows by an order of magnitude as
photometric evidence degrades---and it is the single-render end of that range,
not the thirty-seed average, that describes how synthetic corpora are actually
built.

\section{Discussion}\label{sec:disc}

\textbf{Why the ordering inverts.} The radiance integrand at a surface point
$P$ seen from direction $\boldsymbol\omega$ splits as
\begin{equation}
f(P;\boldsymbol\omega)=f_{\mathrm{ind}}(P)+f_{\mathrm{dep}}(P,\boldsymbol\omega),
\label{eq:decomp}
\end{equation}
with $f_{\mathrm{ind}}$ collecting view-independent contributions---shadowing
from area lights, indirect illumination, caustics projected onto diffuse
surfaces---and $f_{\mathrm{dep}}$ the Fresnel, glossy and refractive
terms~\cite{veach1997robust,pharr2023pbrt}. Under MC integration the variance
field inherits the split. After warping, both pixels sample the same $P$, so
the fields agree to the extent $f_{\mathrm{ind}}$ dominates. In Lambertian
regions it dominates almost entirely, and the warped fields align down to the
estimator noise floor; specular and refractive materials are governed by terms
that differ between the views even under perfect geometric alignment. The
intuition that complex transparent materials should carry more cross-view
structure has the ordering backwards. A second mechanism plausibly also acts
on glass---it is the heavy-tailed regime where a sample-variance estimator is
itself least stable---and our data does not cleanly separate the two.

What makes the decomposition more than a story fitted after the fact is that
the ordering it predicts appears at four levels of evidence that share no
machinery (Table~\ref{tab:material}). It holds in the raw correlation; it
survives a change of renderer; and it reappears \emph{inside} the learned
probe, which was never told what a material is---the probe extracts the cue
where $f_{\mathrm{ind}}$ dominates and sits at chance on glass, in the isolated
single-render case exactly so. A confound would have to reproduce the same
ordering in a correlation coefficient, in a second light-transport
implementation and in the internal behaviour of a network trained on patches.
The integrand split does so by construction.

\begin{table}[t]
\caption{The material ordering recurs at four independent levels of evidence.
Rows 1--2 are correlations; rows 3--4 are 2AFC accuracies of the learned probe,
where chance is $0.500$.}
\label{tab:material}
\centering
\setlength{\tabcolsep}{4pt}
\begin{tabular}{llcc}
\toprule
Level of evidence & Quantity & Non-glass & Glass \\
\midrule
Mitsuba, 20 scenes  & $\rhow$        & $0.779$ & $0.301$ \\
Cycles, 5 scenes    & $\rhow$        & $0.726$ & $0.487$ \\
Probe, $\sigma^2$ input & 2AFC       & $0.793$ & $0.526$ \\
Probe, residual $N{=}1$ & 2AFC       & $0.549$ & $0.492$ \\
\bottomrule
\end{tabular}
\end{table}

\textbf{What a real sensor would give instead.} We could not measure the
warped correlation on a RAW binocular capture: consumer ISP denoising and tone
mapping destroy exactly the statistics in question. We therefore state the
contrast in the form the sensor literature licenses.

For a linear imager the noise variance of a pixel is an affine function of
that pixel's own mean signal, $\sigma^2 = K\mu + \sigma^2_{\mathrm{dark}}$:
this is the photon-transfer relation on which the EMVA~1288 characterisation
standard is built~\cite{emva1288}, and the same Poissonian--Gaussian form is
the accepted model for raw sensor data in vision~\cite{foi2008poissonian}.
Two consequences follow, and the second is the one that matters.

First, it is \emph{not} the case that a real stereo pair has no cross-view
variance structure. Corresponding pixels image the same surface point and have
similar intensities---that is what photometric matching exploits---so if
variance tracks intensity, the warped variance fields of a real pair must be
similar too. A na\"ive $\rhow$ on real data would not be near zero, and
reporting one would prove nothing.

Second, and decisively: that structure is intensity restated. It is exactly
the information a photometric matcher already has, contributing nothing
beyond it. MC variance is different in kind---set by how hard the pixel is to
integrate rather than by how bright it is, per Eq.~\eqref{eq:decomp}---so its
cross-view alignment is information the photometric channel does not carry.
The two regimes differ not in whether a warped variance correlation exists,
but in whether it is redundant. A reliance learned on the rendered structure
is therefore miscalibrated evidence on real data.

\textbf{Mitigations, one of which is already deployed.} Raising spp shrinks
the residual but leaves its cross-view structure intact
(Sec.~\ref{sec:field}), so it is at best partial. Coupling the two views' seeds is worse
than doing nothing: we measured same-stream residual correlation of $0.45$ at
identical pixel coordinates, which injects a sample-level dependence strictly
stronger than the field-level one it would be meant to remove.

Denoising removes the residual by construction, and it is common
practice---Yan \emph{et al.}~\cite{yan2025whatmakes} apply OptiX denoising at
1024 spp for a $7.5\times$ speed-up---but it is not a settled default, since
the same study finds render noise helps on KITTI. The switch deciding whether
this cue is present is therefore already being thrown in both directions, on
grounds that concern noise magnitude alone: a corpus built at one quality tier
may carry a cross-view structure that the same generator does not produce at
another.

\textbf{Limitations.} The probe establishes learnability, not the reliance of
any deployed architecture. That measurement cannot be made on an existing
checkpoint, because the corpora today's models were pre-trained on carry no MC
residual to rely on; it requires pre-training at scale on two path-traced
corpora identical in geometry, materials and lighting and differing only in
whether the residual is cross-view aligned. That is a dataset-construction
project rather than an ablation, and we would rather specify it than
approximate it. The warp
assumes rectified stereo with known ground-truth disparity, which holds by
construction for synthetic data. Two
renderers is better than one but is not a survey. And the real-sensor contrast
remains an argument from the sensor noise model rather than a measurement: the
decisive experiment is not a na\"ive $\rhow$ on real data, which the model says
would be large and uninformative, but whether the correlation survives
conditioning on the two intensities. Running that on real captures requires
repeated RAW exposures of a static scene from a calibrated stereo pair---a
combination the existing corpora do not provide, since RAW burst datasets are
monocular and rectified stereo benchmarks ship one processed shot per view.

\section{Conclusion}\label{sec:concl}

Independence at the sample level does not buy independence at the level a
network consumes. The two claims coexist in path-traced stereo, and the gap
between them is a matching cue: deterministic, renderer-generic, learnable,
strongest exactly where the variance itself is smallest, and---because a real
sensor's variance is dictated by its own signal---with no non-redundant
counterpart in a binocular capture.

Three questions remain, none settleable by characterisation alone: whether
deployed architectures lean on the cue when trained at scale, what it costs
them on real data, and whether denoising and its alternatives remove it
without removing the data's utility. We have tried to leave each well-posed
rather than merely open.

\bibliographystyle{IEEEtran}  
\bibliography{references}

\end{document}